\PassOptionsToPackage{numbers}{natbib}
\documentclass{article}
\usepackage[preprint]{neurips_2025}
\usepackage[utf8]{inputenc} 
\usepackage[T1]{fontenc}    
\usepackage{hyperref}       
\usepackage{url}            
\usepackage{booktabs}       
\usepackage{amsfonts}       
\usepackage{nicefrac}       
\usepackage{microtype}      
\usepackage{xcolor}         
\usepackage{amsmath}        
\usepackage{array}          
\usepackage{longtable}      
\usepackage{enumitem}       
\usepackage{graphicx}       
\usepackage{subfigure}

\title{Aligning Brain Signals with Multimodal Speech and Vision Embeddings}

\author{
  Kateryna Shapovalenko\thanks{Equal contribution.}\textsuperscript{1}, 
  Quentin Auster\footnotemark[1]\textsuperscript{1} \\
  \textsuperscript{1}Carnegie Mellon University, Pittsburgh, PA 15213 \\
  \texttt{\{kshapova, qja\}@alumni.cmu.edu}
}

\begin{document}
\maketitle

\begin{abstract}
When we hear the word “house”, we don’t just process sound, we imagine walls, doors, memories. The brain builds meaning through layers, moving from raw acoustics to rich, multimodal associations. Inspired by this, we build on recent work from Meta that aligned EEG signals with averaged \verb|wav2vec2| speech embeddings, and ask a deeper question: which layers of pre-trained models best reflect this layered processing in the brain? We compare embeddings from two models: \verb|wav2vec2|, which encodes sound into language, and \verb|CLIP|, which maps words to images. Using EEG recorded during natural speech perception, we evaluate how these embeddings align with brain activity using ridge regression and contrastive decoding. We test three strategies: individual layers, progressive concatenation, and progressive summation. The findings suggest that combining multimodal, layer-aware representations may bring us closer to decoding how the brain understands language, not just as sound, but as experience.
\end{abstract}


\section{Introduction}

Human auditory perception is a complex, hierarchical process that begins with the detection of acoustic signals and culminates in the comprehension of spoken language. As sound travels from the ears through auditory pathways, it is incrementally transformed into increasingly abstract representations (phonemes, words, and ultimately meaning), distributed across multiple brain regions \cite{holt2022}.

These transformations are not purely acoustic. When we hear language, we may also visualize objects, recall past experiences, or imagine scenes. The brain constructs meaning through a rich interplay of sensory and associative representations, reflecting the inherently multimodal nature of language understanding.

Given this close relationship between auditory stimuli, mental imagery, and neural responses, decoding language from brain signals remains a compelling goal for both neuroscience and artificial intelligence. Prior work by D\'efossez et al.~\cite{defossez2023decoding} showed that embeddings from a pre-trained speech model (\verb|wav2vec2|) could be contrastively aligned with EEG signals, using the average of its final encoder layers.

In this work, we ask a deeper question: which layers of pretrained models best align with neural activity during speech perception? Inspired by the brain’s layered processing, we move beyond averaged representations to perform layer-wise alignment using embeddings from both \verb|wav2vec2| and \verb|CLIP|, the latter offering a lens into the visual associations that may arise when we hear language. We systematically compare three aggregation strategies to evaluate which best aligns with EEG signals recorded while participants listened to a chapter of Alice in Wonderland \cite{brennan_2019}.


\section{Literature Review}

D\'efossez et al.~\cite{defossez2023decoding} aligned EEG signals with audio representations from a pre-trained \verb|wav2vec2| model using a contrastive CLIP-style loss. Their model, built with convolutional and transformer layers, used the average of the final four encoder layers for alignment. While they demonstrated above-chance decoding from EEG, performance was notably higher with MEG data, suggesting limitations in EEG signal quality or modeling.

Our prior work~\cite{auster2025penny} improved EEG decoding by introducing subject-specific attention, personalized spatial mechanisms, and a dual-path RNN, reducing word error rate by up to 1.9\%. These results highlighted the importance of architecture design and subject adaptation in brain-to-speech tasks.

In this study, we take a complementary approach: instead of modifying the decoder, we investigate whether the choice of embedding layer affects alignment quality. We compare representations across depths and modalities using embeddings from \verb|wav2vec2| and \verb|CLIP|, aiming to understand how brain signals reflect hierarchical and multimodal structure in language models.


\section{Dataset}

We use the EEG dataset introduced by Brennan and Hale \cite{brennan_2019}, which includes:

\begin{itemize}
    \item \textbf{EEG data:} 62-channel EEG recordings from 33 participants, totaling approximately 6.7 hours of listening data. The authors excluded 16 participants due to noisy recordings or low comprehension.
    \item \textbf{Audio data:} The audio consists of a recording of Chapter One from Alice in Wonderland, segmented into 12 WAV files.
\end{itemize}

\begin{figure}[ht]
    \centering
    \subfigure[EEG Channel Removal]{%
        \includegraphics[width=0.95\textwidth]{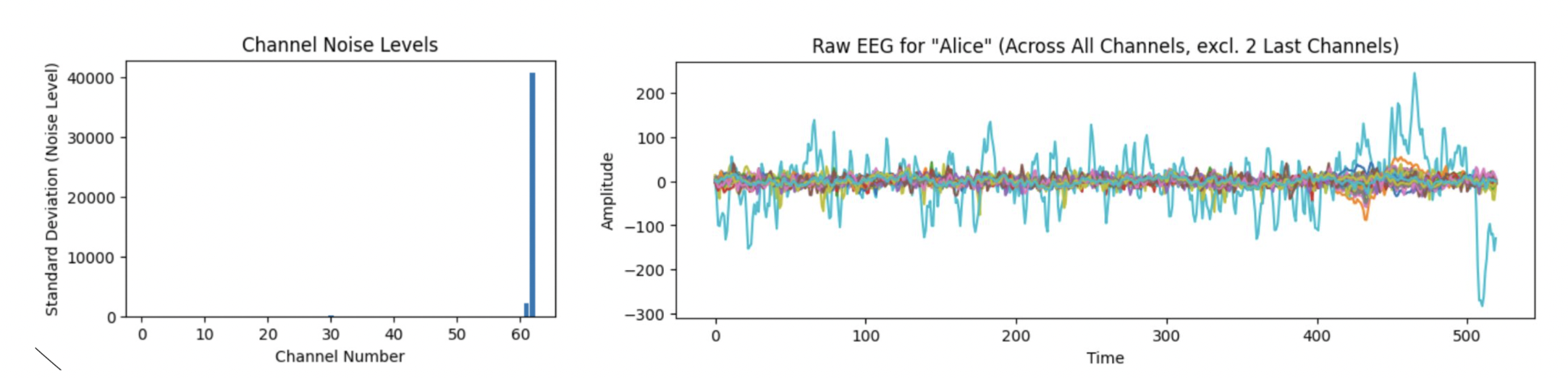}
        \label{fig:pre-proc1}}
    \subfigure[Time- and Frequency-domain Feature Extraction]{%
        \includegraphics[width=0.95\textwidth]{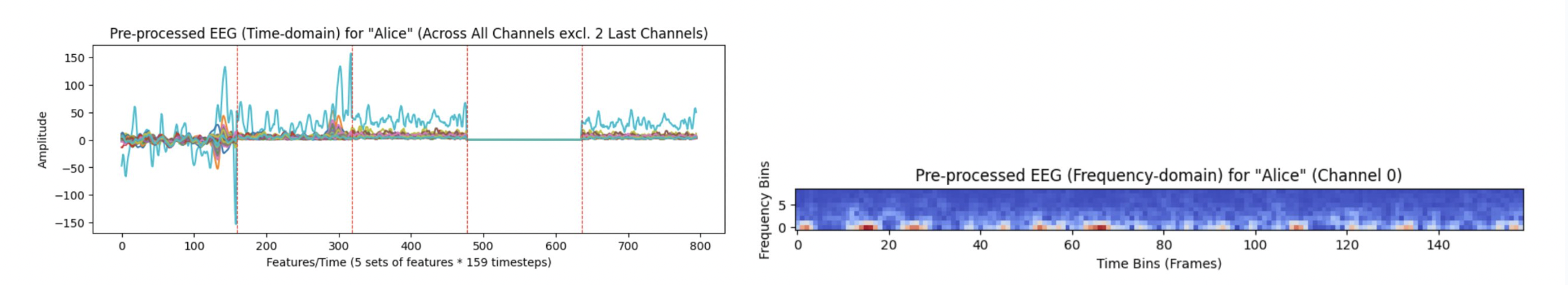}
        \label{fig:pre-proc2}}
    \subfigure[Additional Pre-processing]{%
        \includegraphics[width=0.5\textwidth]{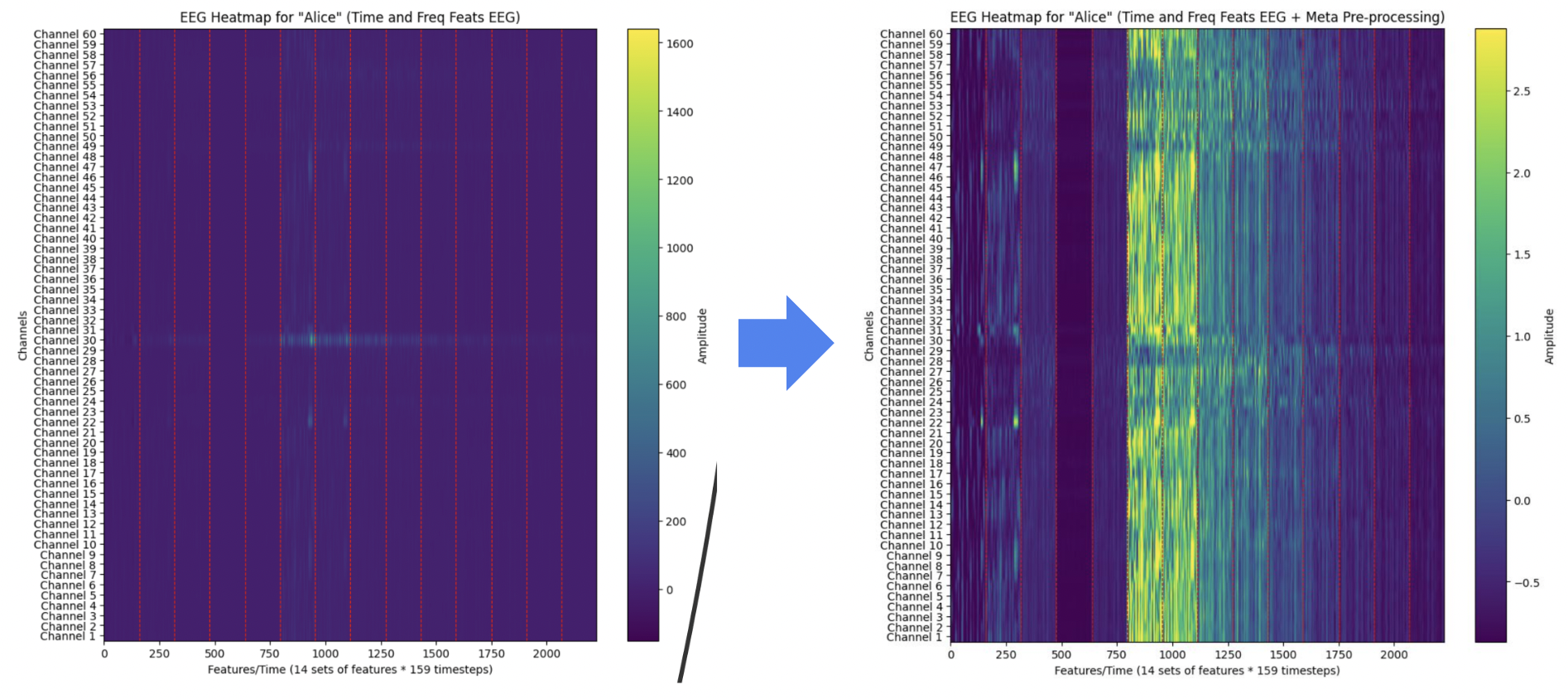}
        \label{fig:pre-proc3}}
    \caption{EEG Data Preprocessing Pipeline: (a) noisy channel removal, (b) time/frequency feature extraction, (c) normalization and outlier correction.}
    \label{fig:preproc-all}
\end{figure}

\section{Data Pre-processing}

We segmented the EEG recordings into word-level chunks using alignment timestamps provided in the original preprocessing pipeline \cite{brennan_2019}. For each word, we extended the segment window by 150 milliseconds before and after the onset to account for neural latency.

To clean the signal, we applied notch filtering around 60 Hz and its harmonics to remove line noise, followed by a high-pass filter at 2 Hz. We removed two noisy channels (VEOG and AUD), which consistently showed high noise levels (see Figure~\ref{fig:pre-proc1}).

We extracted both time- and frequency-domain features. For the time domain, we computed five feature types: (1) mean of the smoothed signal, (2) root mean square (RMS) of the smoothed signal, (3) RMS of the signal envelope, (4) zero-crossing rate, and (5) mean of the envelope. For the frequency domain, we computed nine frequency-bin features using the Short-Time Fourier Transform (STFT). All features were concatenated into a final tensor of shape \verb|[60 channels, 14 features, 159 time frames]| (see Figure~\ref{fig:pre-proc2}).

Following D\'efossez et al., we applied additional preprocessing steps: baseline correction, robust scaling, outlier clipping (5th–95th percentiles), clamping extreme values (above 20 standard deviations), and standard normalization. These steps improved the diversity and stability of extracted features (see Figure~\ref{fig:pre-proc3}).


\section{Methodology}

\begin{figure}[ht]
    \centering
    \includegraphics[width=0.8\textwidth]{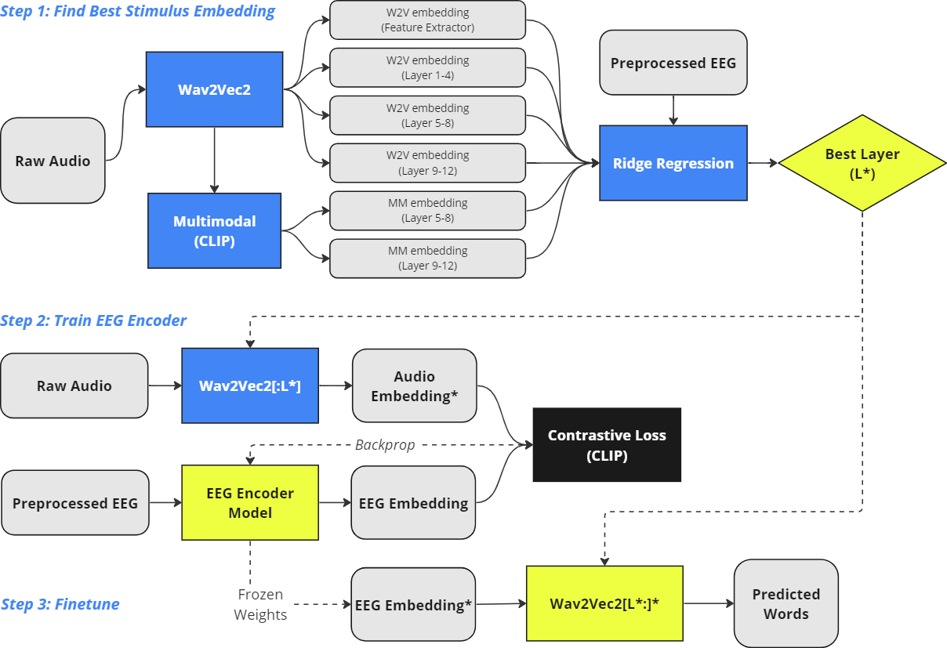}
    \caption{Overview of Methodology.}
    \label{fig:method}
\end{figure}

As shown in Figure~\ref{fig:method}, our approach consists of three main stages.

First, we perform ridge regressions on low-dimensional representations of audio embeddings extracted from pre-trained models at varying network depths. Specifically, we segment the audio stimulus and pass it through (i) the feature extractor and (ii) grouped encoder layers (1–4, 5–8, 9–12) of a pre-trained \verb|wav2vec2| model. In parallel, we pass the corresponding transcriptions through a multimodal \verb|CLIP| model, which produces word-level embeddings from its text encoder. For each embedding (i.e., stimulus), we reduce its dimensionality using PCA or ICA and regress the resulting principal components (\(X\)) onto preprocessed EEG features (\(Y\)) using ridge regression.

Second, we identify the most predictive layer(s) based on ridge regression performance (using test-set \(R^2\) and correlation). We then truncate the \verb|wav2vec2| model at this optimal layer and use it to generate audio embeddings for training an EEG encoder. This encoder follows the same CNN-transformer architecture as in D\'efossez et al.~\cite{defossez2023decoding}, and is trained to map EEG data to the selected stimulus embedding space.

Third, we propose using the trained EEG encoder in combination with the later layers of \verb|wav2vec2| (i.e., layers after the optimal truncation point) for transfer learning. In this setup, the EEG encoder and truncated \verb|wav2vec2| layers are frozen, while the final \verb|wav2vec2| layers are fine-tuned to predict words. Due to non-convergence issues with the EEG encoder during initial training, we leave this step for future work.

\paragraph{Audio Embeddings}

As the subjects in our dataset listened to a chapter of "Alice in Wonderland", we can imagine that the audio traveled through multiple levels of the human brain, being processed and converted into multiple representations - from basic acoustic features to the actual meaning. We hypothesized that one way to account for this complexity is to use different sets of embeddings for the original audio. To obtain such embeddings, we used two different models: \verb|wav2vec2| and \verb|CLIP|. 

\begin{itemize}
    \item \textbf{Wav2Vec2:} A self-supervised model for audio-to-text tasks. We extracted embeddings from 13 layers: (0) the convolutional feature extractor, and (1–12) the transformer encoder layers. These embeddings span low-level acoustic features to higher-level lexical representations.
    \item \textbf{CLIP:} A multimodal model that aligns language and vision. We used the CLIP text encoder to generate embeddings for audio transcriptions, under the hypothesis that listeners may generate visual mental imagery during story comprehension. Like \verb|wav2vec2|, we extracted 13 embedding layers, including the input projection and transformer blocks.
\end{itemize}

In total, we obtained 26 distinct embeddings per stimulus (13 from \verb|wav2vec2|, 13 from \verb|CLIP|), which we refer to as stimulus embeddings.

\begin{figure}[ht]
    \centering
    \includegraphics[width=1\textwidth]{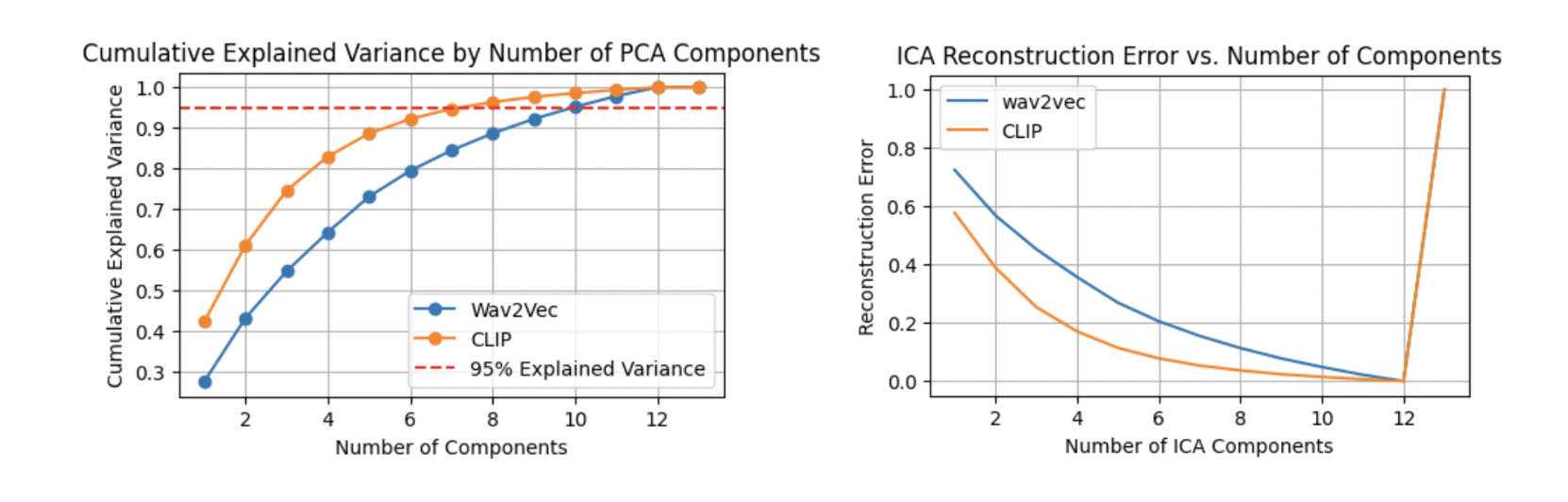}
    \caption{Selecting the Best Number of Components for PCA and ICA.}
    \label{fig:pca_ica}
\end{figure}

\begin{figure}[h]
    \centering
    \includegraphics[width=0.95\textwidth]{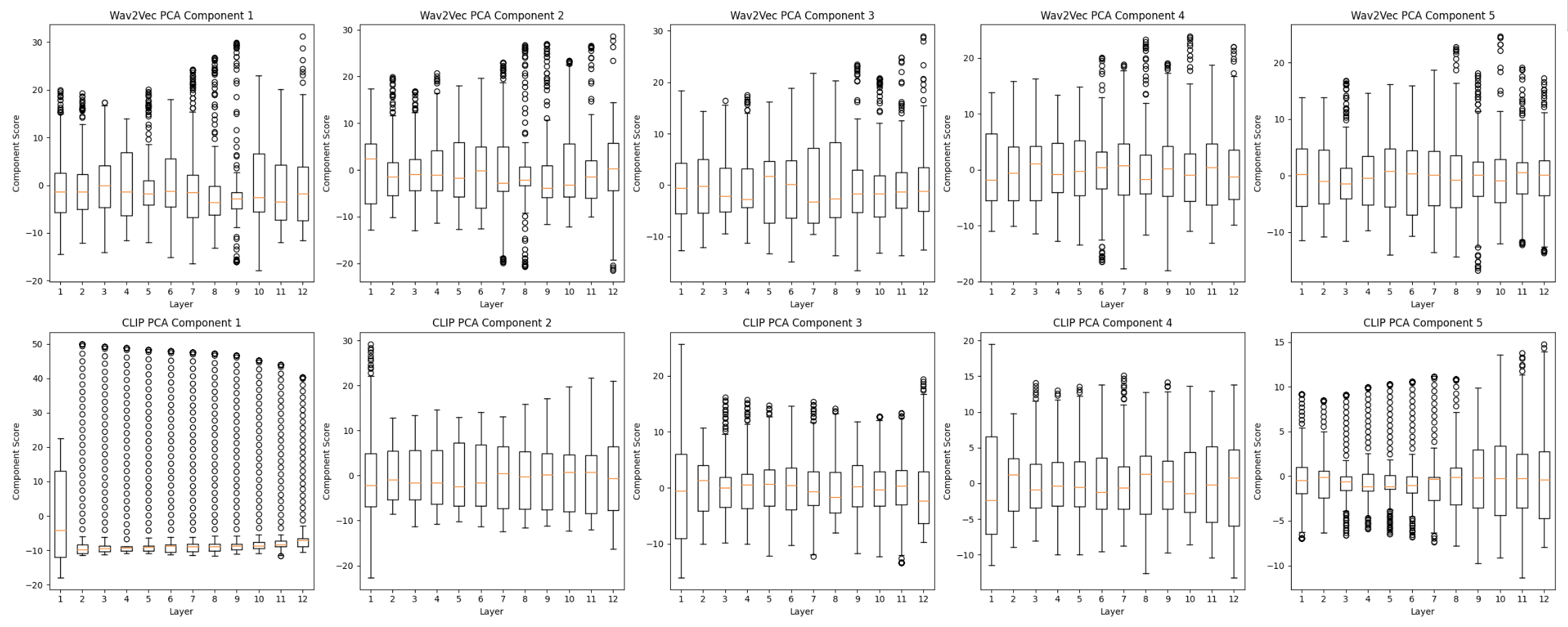}
    \caption{Variance of the First Five PCA Components Across All Stimulus Embeddings}
    \label{fig:pca2}
\end{figure}

To reduce the dimensionality of these embeddings and increase robustness, we applied Principal Component Analysis (PCA) and Independent Component Analysis (ICA). For each word-level stimulus embedding, we selected the top 10 components (see Figures~\ref{fig:pca_ica} and~\ref{fig:pca2}). This reduced the dimensionality from \verb|(13, 122112)| to \verb|(13, 10)| for \verb|wav2vec2|, and from \verb|(13, 81408)| to \verb|(13, 10)| for \verb|CLIP|.

\paragraph{EEG Embeddings} For EEG feature extraction, we used the CNN-transformer encoder architecture described in~\cite{defossez2023decoding}. EEG signals were preprocessed as described earlier and downsampled into fixed-length segments aligned with each word-level audio chunk.

\paragraph{Ridge Regressions}

We investigate whether the layer selection strategy in D\'efossez et al. was optimal. That study averaged the final four layers of \verb|wav2vec2| embeddings. We evaluate whether specific layers (or combinations) offer stronger alignment with EEG. We treat stimulus embeddings as \(X\) (predictors) and preprocessed EEG features as \(Y\) (targets), and compare three regression strategies:

\begin{itemize}
    \item \textbf{Method 1: Single-layer regression.}  
    We train separate ridge regressions using embeddings from each individual layer. This reveals which layer best predicts EEG responses. We conduct this for both PCA and ICA versions of stimulus embeddings. PCA generally outperformed ICA and was used in subsequent methods.
    
    \item \textbf{Method 2: Progressive concatenation.}  
    Embeddings from successive layers are concatenated to form a larger feature vector. This tests whether combining information across layers improves prediction, at the cost of increased dimensionality.
    
    \item \textbf{Method 3: Progressive summation.}  
    Instead of concatenation, embeddings are summed layer-wise. This approach maintains the dimensionality of a single layer and can amplify features consistently present across layers.
\end{itemize}

Each regression was evaluated via cross-validation across multiple regularization parameters (\(\alpha\)). We conducted experiments on EEG data from the top-performing subjects (S04, S13, S19) based on their comprehension scores from post-task questionnaires. For each subject, the data was split 80/20 for training and testing. We plan to extend this analysis to the full dataset in future work.

\section{Training Setup}

We trained the EEG encoder using a CLIP-style contrastive loss. To do so, we fed preprocessed 2D EEG inputs of shape $B \times 60 \times (14 \times 159)$ into an EEG encoder based on the original design from Meta. We then used embeddings obtained by passing raw audio through \verb|wav2vec2|, truncated at varying layer depths.

We trained the model with a batch size of 64 using the \verb|AdamW| optimizer, an initial learning rate of $4 \times 10^{-4}$, and a learning rate scheduler that reduced the learning rate by a factor of 0.5 after four epochs of no improvement.


\section{Results and Discussion}

\subsection{Single-Layer Regression Results}

\begin{figure}[ht]
    \centering
    \subfigure[Regression performance (R² and correlation).]{%
        \includegraphics[width=0.75\textwidth]{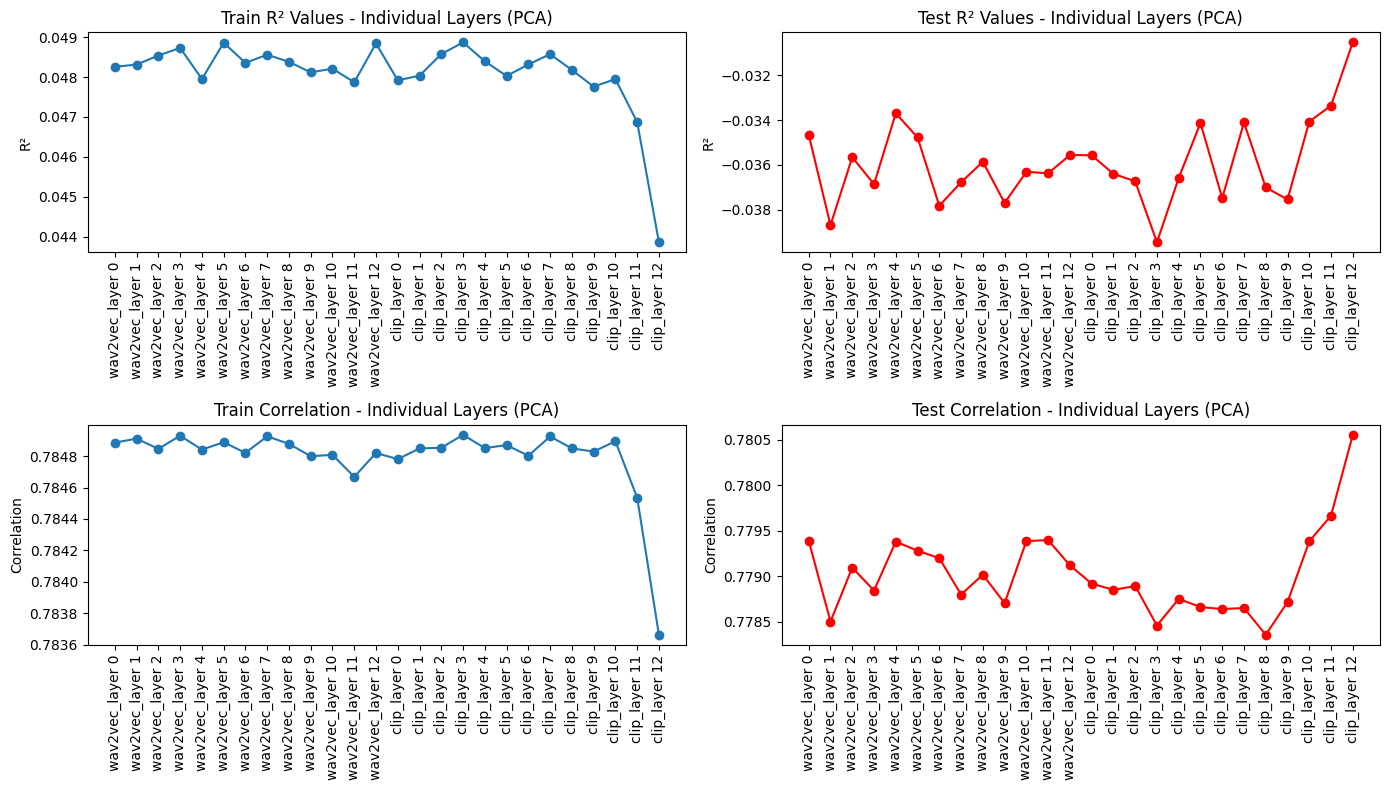}
        \label{fig:reg-ind-pca}
    }\\[1ex]  
    \subfigure[Topographic maps of regression weights.]{%
        \includegraphics[width=0.9\textwidth]{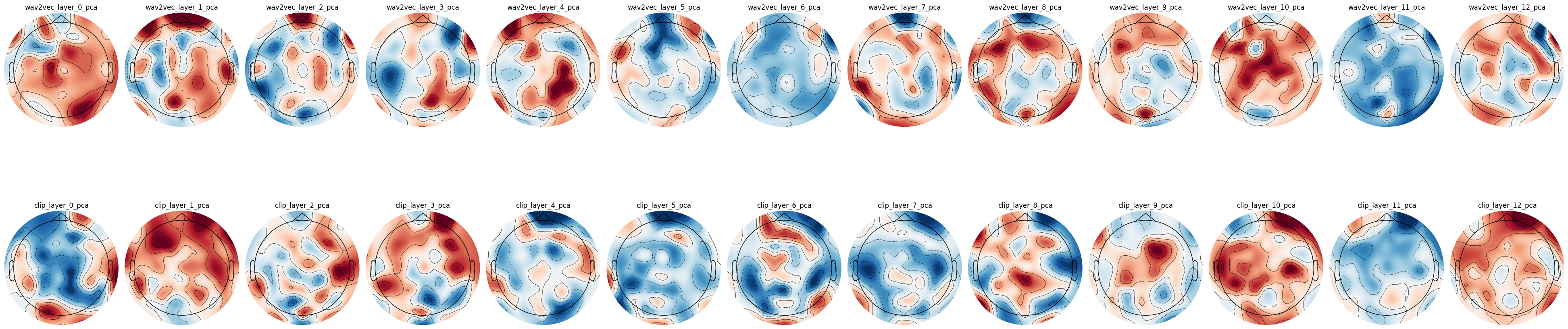}
        \label{fig:reg-ind-pca-brain}}
    \caption{Single-layer regression results (PCA, Subject S04).}
    \label{fig:method1}
\end{figure}

Using PCA-transformed embeddings, we observe modest to negligible predictive power for EEG responses across both the \verb|wav2vec2| and \verb|CLIP| models. Training $R^2$ values are consistently low across all layers, averaging around 0.048, indicating minimal variance explained. However, training correlations remain uniformly high (approximately 0.784), suggesting the model fits the training data well but may overfit. Test results show negative $R^2$ values across all layers, implying performance worse than a naïve baseline. This points to a failure to generalize, a common challenge in brain decoding tasks. The PCA embeddings may not preserve sufficiently relevant features for robust downstream prediction, or the model may be overly complex given the sample size (see Figure~\ref{fig:reg-ind-pca}).

Topographic maps in Figure~\ref{fig:reg-ind-pca-brain} visualize EEG activation patterns across the scalp. Red regions indicate stronger regression weights, while blue indicates weaker ones. Certain layers, e.g., \verb|wav2vec2_layer_0| and \verb|clip_layer_1|, show more centralized and intense activations, suggesting they may capture more salient or shared features aligned with EEG activity.

In contrast, ICA-transformed embeddings exhibit even lower predictive performance. Training $R^2$ values are lower than for PCA, and test $R^2$ values are dramatically negative, especially for \verb|CLIP| layers, suggesting a deeper disconnect between ICA-derived features and EEG responses. These results are shown in Figures~\ref{fig:reg-ind-ica} and~\ref{fig:reg-ind-ica-brain} in the Appendix.

\begin{figure}[ht]
    \centering
    \subfigure[Regression performance (R² and correlation).]{%
        \includegraphics[width=0.8\textwidth]{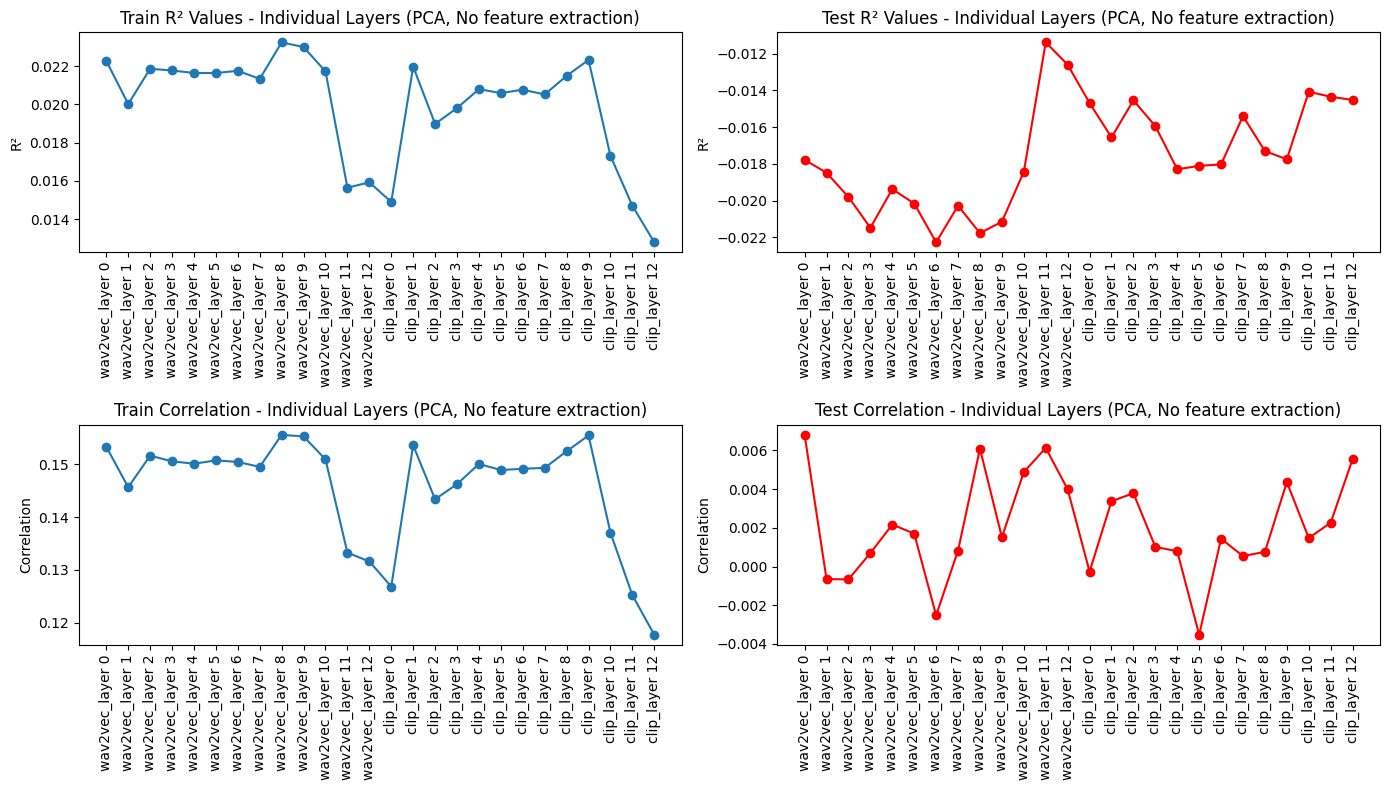}
        \label{fig:reg-ind-pca-no-feats}}\\[1ex]
    \subfigure[Topographic maps of regression weights.]{%
        \includegraphics[width=0.95\textwidth]{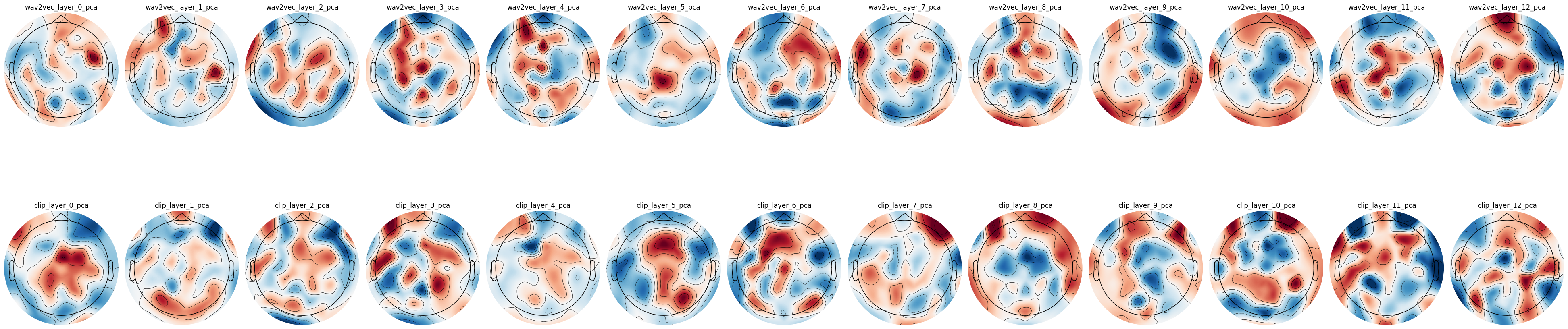}
        \label{fig:reg-ind-pca-no-feats-brain}}
    \caption{Single-layer regression results before feature extraction (PCA, Subject S04).}
    \label{fig:method1-pca-no-feats}
\end{figure}

To assess robustness, we repeated the same analyses using EEG signals extracted earlier in the pipeline (before feature extraction) and expanded the subject pool from one (S04) to three high-comprehension individuals (S04, S13, S19). Results were consistent across these variants, further suggesting that middle layers of both \verb|wav2vec2| and \verb|CLIP| may encode the most relevant information for EEG prediction (see Figures~\ref{fig:reg-ind-pca-no-feats} and~\ref{fig:reg-ind-pca-no-feats-brain}).

\subsection{Progressive Layer Aggregation Results - Concatenation vs. Summation}  

We next evaluated two strategies for aggregating embeddings across layers: progressive concatenation and progressive summation. 

\begin{figure}[ht]
    \centering
    \subfigure[Regression performance (R² and correlation).]{%
        \includegraphics[width=0.75\textwidth]{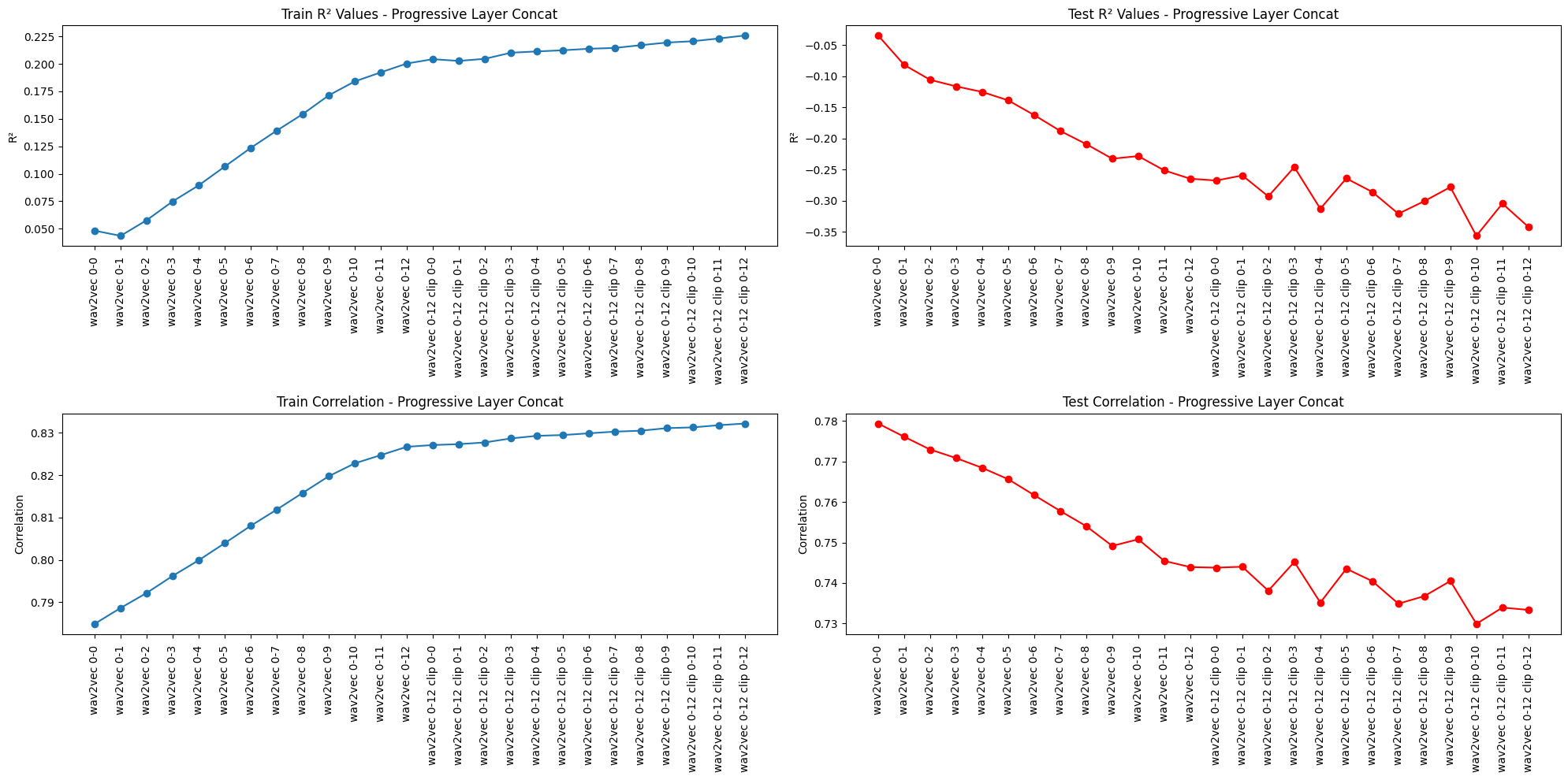}
        \label{fig:reg-concat}
    }\\[1ex]  
    \subfigure[Topographic maps of regression weights.]{%
        \includegraphics[width=0.9\textwidth]{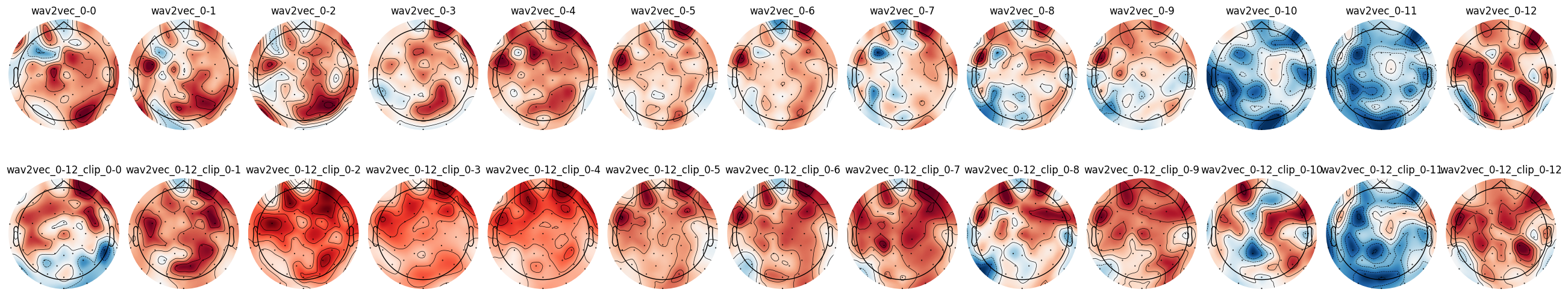}
        \label{fig:reg-concat-brain}}
    \caption{Progressive concatenation results (PCA, Subject S04).}
    \label{fig:method2}
\end{figure}

In the \textbf{progressive concatenation setup}, each successive group of layers was concatenated into a higher-dimensional feature vector. As shown in Figure~\ref{fig:reg-concat}, this strategy steadily improved training performance: the training correlation increased from 0.7849 for \verb|wav2vec2_0-0_pca_concat| to 0.8322 for the full-layer configuration \verb|wav2vec2_0-12_clip_0-12_pca_concat|. Training $R^2$ values followed a similar upward trend, suggesting that combining multiple layers provides a richer stimulus representation for EEG decoding. However, test $R^2$ values steadily decreased from -0.0347 to -0.3421, indicating poor generalization. This is consistent with overfitting: while additional layers capture more variance in training data, they introduce noise or redundancy that hurts performance on unseen samples. Notably, most of the improvement comes from \verb|wav2vec2| layers; adding \verb|CLIP| layers beyond that contributes marginally or not at all.

Topographic maps for the concatenated embeddings (Figure~\ref{fig:reg-concat-brain}) show increased diversity in activation patterns when aggregating \verb|wav2vec2| layers, while \verb|CLIP| layers yield more homogeneous patterns.

\begin{figure}[ht]
    \centering
    \subfigure[Regression performance (R² and correlation).]{%
        \includegraphics[width=0.75\textwidth]{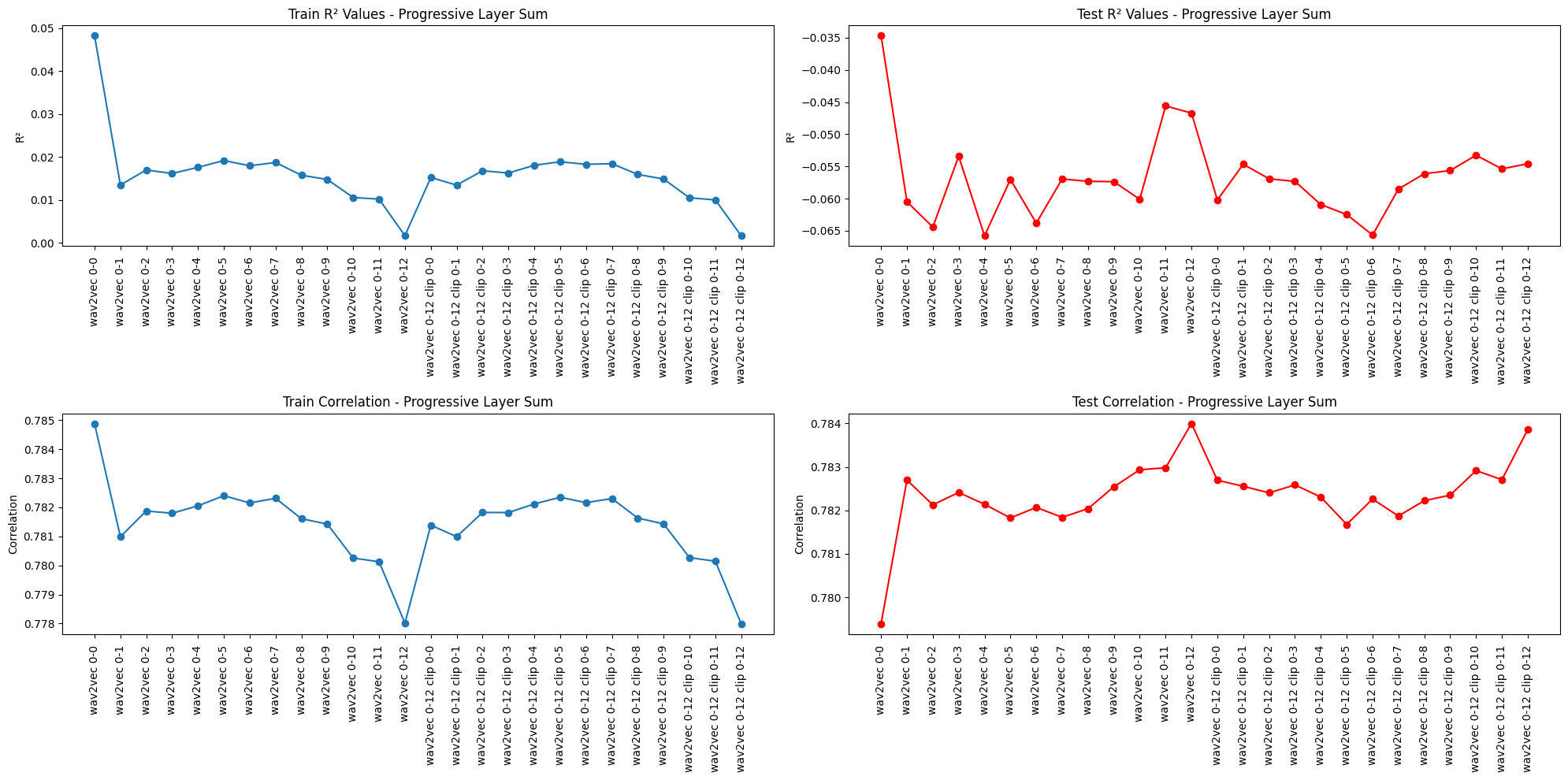}
        \label{fig:reg-sum}
    }\\[1ex]  
    \subfigure[Topographic maps of regression weights.]{%
        \includegraphics[width=0.9\textwidth]{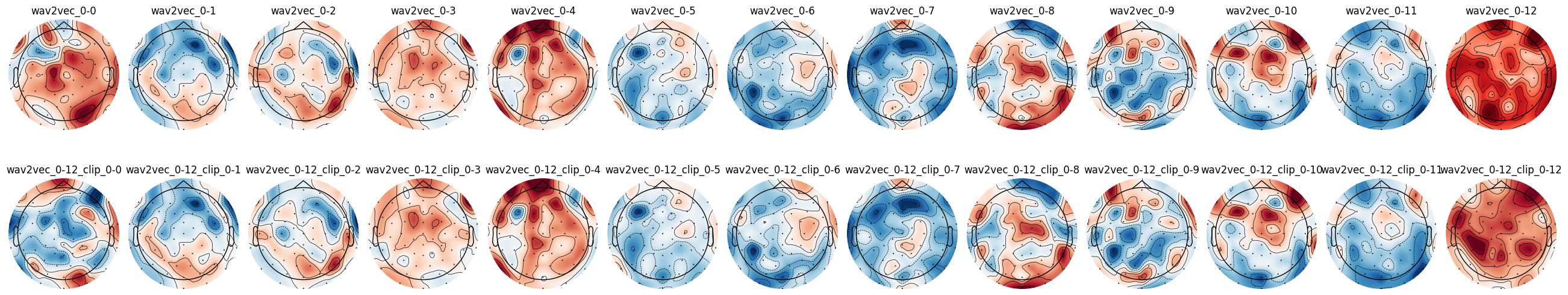}
        \label{fig:reg-sum-brain}}
    \caption{Progressive summation results (PCA, Subject S04).}
    \label{fig:method3}
\end{figure}

In the \textbf{progressive summation setup}, we summed embeddings layer by layer, preserving dimensionality and emphasizing shared features. As seen in Figure~\ref{fig:reg-sum}, both training $R^2$ and correlation initially dipped after layer 0 but then increased steadily until around layer 7–8 (e.g., \verb|wav2vec2_0-7_pca_sum|), after which performance plateaued or declined. This suggests that early-to-mid layers encode features most aligned with EEG representations, while deeper layers may introduce noise or abstract representations not directly reflected in the signal.

Unlike concatenation, summation improved test performance as more layers were added. Test $R^2$ and correlation values increased alongside training metrics, indicating better generalization. Topographic maps (Figure~\ref{fig:reg-sum-brain}) show that the summed layers activated more diverse and distributed brain regions, compared to the flatter profiles seen in deeper concatenated layers.

\subsection{Contrastive Decoding Experiments}

We implemented the contrastive decoding setup, but were unable to achieve convergence across multiple intermediate layer cutoffs. This may be due to training on a single subject, the complexity of the EEG encoder, or limited alignment between the EEG and audio embedding spaces. We plan to investigate these factors in future work.


\section{Conclusions}

This study examined how different layers of pre-trained audio models align with EEG signals recorded during natural speech perception. We found that mid-sequence layers of both \verb|wav2vec2| and \verb|CLIP| provided the most consistent alignment, suggesting that these stages capture a meaningful balance between low-level acoustic and high-level linguistic features.

Despite achieving strong correlations on training data, our models struggled to generalize, indicated by negative test $R^2$ values across all conditions. This reflects the broader challenge of overfitting in EEG decoding and highlights the difficulty of modeling shared neural patterns across individuals.

Among our tested aggregation strategies, progressive summation proved more robust than concatenation, especially on test data. However, even this method failed to fully generalize, pointing to fundamental limitations in current embedding spaces and decoding architectures. Bridging this gap remains an open challenge for brain-to-audio alignment models.

\section{Future Work}

Future work should explore subject-invariant architectures and larger multi-subject datasets to enhance generalization. Also, further analysis of alternative embedding spaces may also improve alignment between EEG and audio features.


\section*{Acknowledgment}
We would like to thank Professor Leila Wehbe of Carnegie Mellon University for her guidance and support throughout this project.

This work was conducted as part of the Carnegie Mellon University course 10-733 Representation and Generation in Neuroscience and AI (Spring 2024): \url{https://www.cs.cmu.edu/~lwehbe/10733_S24/}


\bibliography{refs}

@article{defossez2023decoding,
    title={Decoding speech perception from non-invasive brain recordings},
    author={
        D{\'e}fossez, Alexandre 
        and Caucheteux, Charlotte 
        and Rapin, J{\'e}r{\'e}my 
        and Kabeli, Ori 
        AND King, Jean-R{\'e}mi
    },
    journal={Nature Machine Intelligence},
    year={2023},
    volume = {5},
    pages = {1097--1107},
    DOI = {https://doi.org/10.1038/s42256-023-00714-5}
}

@article{brennan_2019,
    author = {J.R. Brennan and J.T. Hale},
    title = "Hierarchical structure guides rapid linguistic predictions during naturalistic listening",
    journal = "PLoS ONE",
    volume = "14",
    number = "1",
    year = "2019",
    DOI = "https://doi.org/10.7302/746w-g237"
}

@book{holt2022,
    author = {Lori L. Holt · Jonathan E. Peelle Allison B. Coffin · Arthur N. Popper Richard R. Fay},
    title = {Speech Perception},
    publisher = {Springer},
    year = {2022}
}

@misc{auster2025penny,
  title        = {A Penny for Your Thoughts: Decoding Speech from Inexpensive Brain Signals},
  author       = {Quentin Auster and Kateryna Shapovalenko and Chuang Ma and Demaio Sun},
  year         = {2025},
  eprint       = {2511.04691},
  archivePrefix= {arXiv},
  primaryClass = {cs.SD},
  url          = {https://arxiv.org/abs/2511.04691}
}


\pagebreak
\section*{Annexes}

\begin{figure}[ht]
    \centering
    \subfigure[Regression performance (R² and correlation).]{%
        \includegraphics[width=0.8\textwidth]{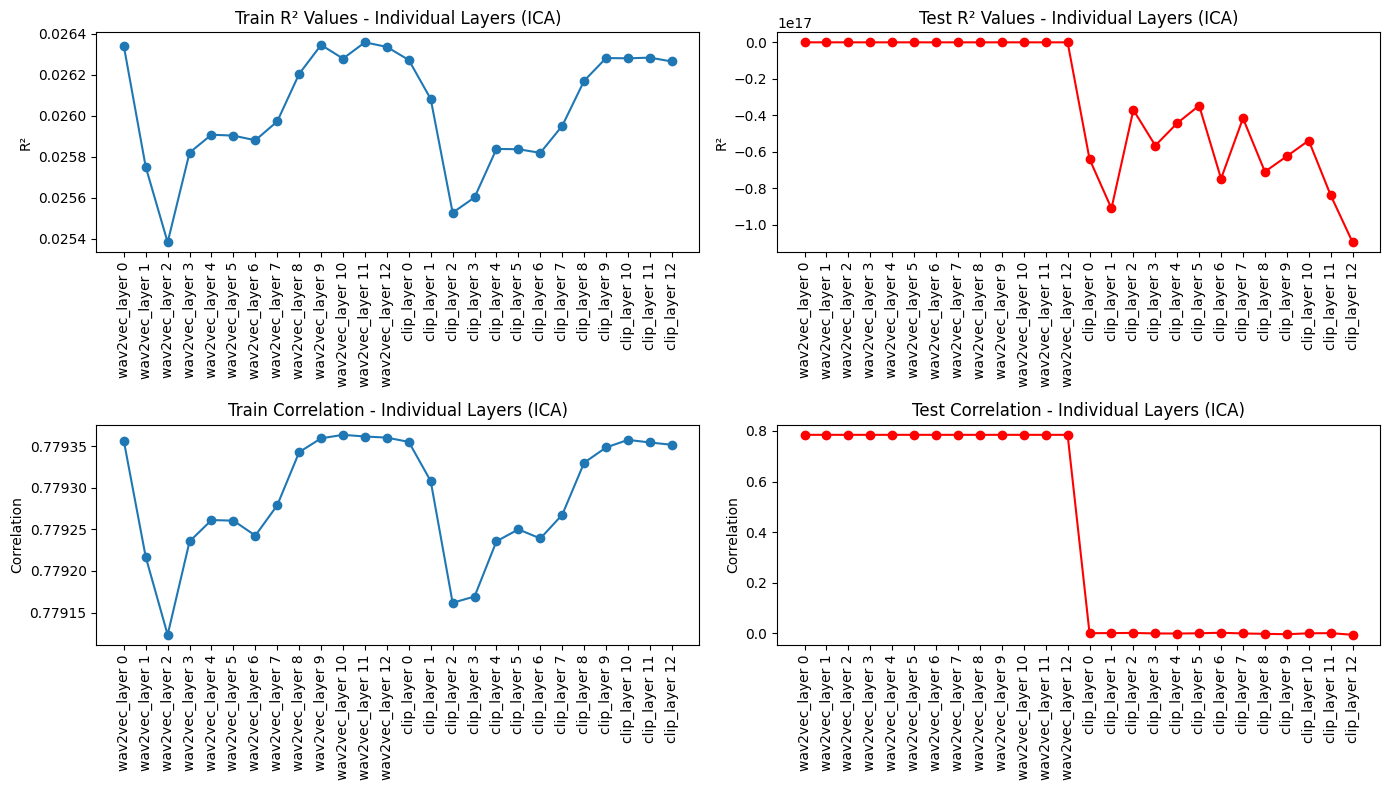}
        \label{fig:reg-ind-ica}}\\[1ex]
    \subfigure[Topographic maps of regression weights.]{%
        \includegraphics[width=0.95\textwidth]{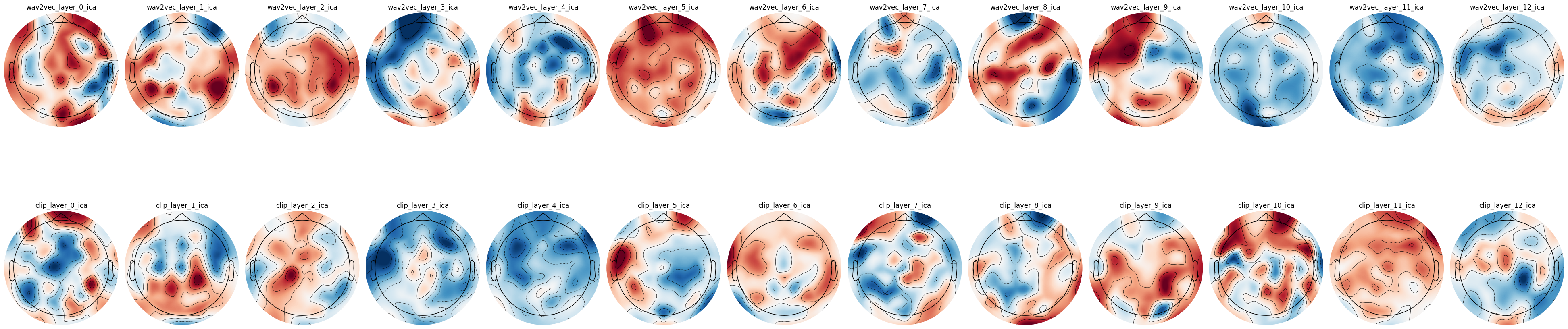}
        \label{fig:reg-ind-ica-brain}}
    \caption{Single-layer regression results (ICA, Subject S04).}
    \label{fig:method1-ica}
\end{figure}

\pagebreak

\end{document}